# Subject-Independent Drowsiness Recognition from Single-Channel EEG with an Interpretable CNN-LSTM model

Jian Cui
*Fraunhofer Singapore*
*Nanyang Technological University*
Singapore
cuijian@ntu.edu.sg

Zirui Lan
*Fraunhofer Singapore*
Singapore
LANZ0001@e.ntu.edu.sg

Tianhu Zheng
*Nanyang Technological University*
Singapore
ZHEN0129@e.ntu.edu.sg

Yisi Liu
*Fraunhofer Singapore*
Singapore
liu.yisi@fraunhofer.sg

Olga Sourina
*Fraunhofer Singapore*
*Nanyang Technological University*
Singapore
EOSourina@ntu.edu.sg

Lipo Wang
*Nanyang Technological University*
Singapore
ELPWang@ntu.edu.sg

Wolfgang Müller-Wittig
*Fraunhofer Singapore*
*Nanyang Technological University*
Singapore
Wolfgang.Mueller-wittig@fraunhofer.sg

*Abstract*—For EEG-based drowsiness recognition, it is desirable to use subject-independent recognition since conducting calibration on each subject is time-consuming. In this paper, we propose a novel Convolutional Neural Network (CNN)-Long Short-Term Memory (LSTM) model for subject-independent drowsiness recognition from single-channel EEG signals. Different from existing deep learning models that are mostly treated as black-box classifiers, the proposed model can "explain" its decisions for each input sample by revealing which parts of the sample contain important features identified by the model for classification. This is achieved by a visualization technique by taking advantage of the hidden states output by the LSTM layer. Results show that the model achieves an average accuracy of 72.97% on 11 subjects for leave-one-out subject-independent drowsiness recognition on a public dataset, which is higher than the conventional baseline methods of 55.42%-69.27%, and state-of-the-art deep learning methods. Visualization results show that the model has discovered meaningful patterns of EEG signals related to different mental states across different subjects.

Code of the paper is available from: https://github.com/cuijiancorbin/Subject-Independent-Drowsiness-Recognition-from-Single-Channel-EEG-with-an-Interpretable-CNN-LSTM

I. INTRODUCTION

Resulted from sleep deprivation or a prolonged period of monotonous driving, drowsiness not only can cause inefficiency of work but has also become a major factor of car accidents. Electroencephalography (EEG) is one of the best techniques to detect drowsiness, which measures voltage fluctuations on the scalp that reflect mental activities occurring in the brain. The common practice of using EEG signals for mental state recognition is training a subject-specific classifier with EEG signals captured from the subject under different mental states. However, it is inconvenient to use such subject-dependent methods for drowsiness recognition, since induction of drowsiness could be long and tedious. The alternative way is building a subject-independent drowsiness recognition system, which uses EEG signals from other subjects to train the classifier. However, achieving a satisfactory accuracy with the method is still challenging, since there is a large variability of EEG signals among different subjects and the noise and artifacts contained in the data could also affect the recognition results. This motivates us to find out 1) what common features of EEG signals shared by different subjects can be used to distinguish between the alert and drowsy states, and 2) to which extent the noise and artefacts contained in the signals affect the recognition accuracy.

Recent progress in the field of EEG signal processing has introduced deep learning methods for mental state recognition from EEG signals. However, existing models are mostly treated as black-box classifiers while what has been learned by the models are underexplored. In fact, visualization on what are learned by the model is an important process for validation, without which the work on improving the model towards higher accuracy become trial-and-error.

In this paper, we propose an interpretable deep learning model for subject-independent drowsiness recognition from EEG signals. The model is designed to process single-channel EEG signals, considering consumer-grade EEG devices with one or multiple electrodes have advantages over medical-grade ones in the aspects of ease of set-up and low price to be used on daily basis. In the model structure, Convolutional Neural Network (CNN) is used to extract a sequence of features, which are linked to classification with a Long Short-Term Memory (LSTM) layer. In order to enable the model to "explain" its decisions, we have designed a novel visualization technique that can reveal local regions of the input signal containing important features for the classification based on the hidden states output by the LSTM model. In the following part of the paper, related work is reviewed in Section II. Materials and methods are illustrated in Section III. The model is evaluated in Section IV, which is followed by discussion and future work in Section V and conclusion Section VI, respectively.

## II. RELATED WORK

As an affordable brain imaging technique, EEG uses a set of electrodes to measure voltage fluctuations on the head resulted from ionic current in the cerebral cortex. It has been revealed by many studies that there exists a strong relationship between drowsiness and the oscillation patterns of EEG signals [1-3]. Drowsiness recognition from EEG signals with Conventional methods has been extensively studied [4, 5], where features are extracted from the raw data combined with machine learning models for classification.

Recent progress in the field of EEG signal processing has introduced deep learning for classifying different mental states. In comparison to traditional methods, deep learning has a better generalization capability to handle large amount of data over conventional methods in many fields. Among the deep models, CNN and Recurrent Neural Network (RNN) are the most commonly used ones for EEG signal processing [6]. LSTM is a modified version of RNN that has a better capability to handle to the problem of vanishing gradient. These models have been extensively considered for recognition of different mental states from signal-channel EEG signals. Sors et al. [7] proposed a convolutional neural network for prediction of 5-class sleep stages from raw EEG data. Ding et al. [8] built a deep learning model using cascaded CNN and attention mechanism for drivers drowsiness detection from EEG to be used on mobile devices. Fahimi et al. [9] proposed an end-to-end deep CNN to distinguish between attentive and inattentive EEG signals from single-channel EEG data. Akara et al. [10] proposed a deep learning model called DeepSleepNet, which uses CNN to extract features from EEG epochs and use bidirectional-LSTM to learn transition rules for sleep stage recognition. Bresch et al. [11] designed a hybrid model with both CNN and LSTM layers for sleep stage classification. The model achieves an agreement with a human annotator of Cohen's Kappa of $\sim 0.73$ with training data from 19 subjects.

Although existing work exhibits encouraging progress on using deep learning techniques to classify single-channel EEG signals, there is still little insight on what features have been discovered by the deep learning methods to distinguish different mental states. In fact, visualization on what has been learned by the deep learning networks is as important as achieving a high classification accuracy, since it not only has become an essential procedure of model validation but also potentially reveals interesting EEG patterns that are not discovered before. In our previous work [12], we made the initial attempt of designing an interpretable CNN model with a Global Average Pooling (GAP) layer, which enables the Class Activation Map (CAM) method to be used for visualization of the learned shared EEG features across different subjects. However, the GAP layer simply averages the extracted features in the temporal dimension, without considering the valuable temporal dependency information in the data. In this paper, we propose a novel CNN-LSTM model, which uses CNN to extract features and LSTM to learn the temporal dependence of the feature sequence. We also designed a technique to reveal what has been learned by the model to differentiate different mental modes by a novel design of the LSTM layer.

## III. MATERIALS AND METHODS

### A. Data preparation

In this study, we use a public EEG dataset, which is obtained from 27 subjects (aged between 22–28) in a 90-minute sustained-attention virtual driving task [13]. In the experiment, the participants were required to sustain their attention and respond immediately to the randomly introduced lane-departure events, which drifted the car away from the central lane. The occurring time of event onset, response onset, and response offset were recorded and aligned with the timeline of the EEG data. The level of drowsiness can be reflected by the local Reaction Time (local-RT), which is defined by the time interval between the event onset and response onset for each lane-departure event. The samples are labeled with a method proposed by Wei et al. [14]. In the method, the baseline 'alert-RT', which is defined as the $5^{th}$ percentile of the local RTs for one session, is used as a threshold to categorize the samples in each session in order to minimize the individual difference. Another metric called "global-RT" is introduced in order to reflect the subject state in a relatively long period, and it is calculated by averaging the local RTs within a 90-second window before the car drift event. The samples are labeled with 'alert' when both local-RT and global-RT are shorter than 1.5 times alert-RT, while they are labeled with 'drowsy' when both local-RT and global-RT are longer than 2.5 times alert-RT. The method excludes the transitioning period with intermediate RTs since they may be attributed to other unknown processes, e.g., mind-wandering, and the exclusion of transitioning trials could benefit the accuracy of discriminating drowsy trials [14].

The EEG signals were captured using a wired EEG cap containing 32 Ag/AgCl electrodes (30 EEG electrodes and 2 reference electrodes) based on a modified international 10–20 system. The pre-processed version of the dataset [15] is used, where the EEG data digitalized at 500 Hz were processed with 1-Hz high-pass and 50-Hz low-pass filters followed by artefact rejection. We use the EEG data from the Oz channel in this study, since the channel was found to contain distinctive features with least noise for distinguishing between drowsy and alert EEG signals [16]. The EEG signals are down-sampled to 128 Hz. We extract a 3-second length sample prior to each car deviation event forming a vector of size 1 (channel) × 384 (sampling points).

We further balance the data for each class and each subject by taking the following steps: 1) sessions with less than 50 samples of either class are discarded; 2) if there are multiple sessions belonging to the same subject, the one with the most balanced class distribution is used; 3) samples from each session are further balanced by choosing the most representative ones with shortest (for alert class) or longest (for fatigue class) local RTs from the majority class. The first and third steps ensure that the classifiers are trained with samples from balanced classes, while the second step balances data from different subjects so that the classifier will not favor prediction of a specific subject.

By taking the steps above, we have built a compact dataset that contains the most representative alert and drowsy EEG samples from different subjects for the study. The dataset contains 2022 samples from 11 different subjects, and the number of samples for each subject is shown in TABLE I. A more detailed description on the extraction process can be found from our previous paper [12]. The extracted dataset has been uploaded online, which is accessible from [17].

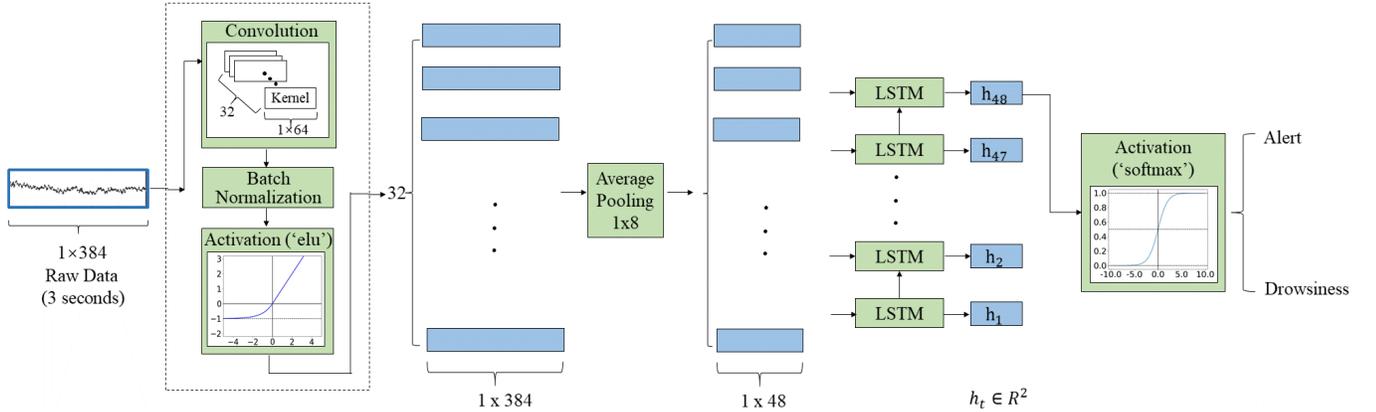

Figure 1. structure of the convolutional LSTM model.

### B. Design of network and visualization technique

#### 1) Core idea

RNNs are a class of neural networks that are capable of handling sequential data. As a special kind of RNN, LSTM unit [18] was introduced to capture the long and short-term dependencies of the sequence for solving the problem of vanishing gradient. The LSTM model is actually a function that sequentially maps a set of input elements $X = (x_1, x_2, ..., x_T)$ to a set of hidden feature-state representations $(h_1, h_2, ..., h_T)$, $h_i \in \mathbb{R}^D$, where $D$ is the dimension of each representation. The learning process can be described as

$$h_t \leftarrow LSTM(x_t, h_{t-1}) \qquad (1)$$

The input elements $X = (x_1, x_2, ..., x_T)$ sent to the LSTM model are sequential representations of the data, e.g., word vectors in a sentence. For our case of EEG signal processing, $X$ is modeled as a sequence of features extracted from an EEG sample $S = (s_1, s_2, ..., s_n)$, where each $s_i$ is a sampling point of the signal and $n$ is the length of the signal.

TABLE I. NUMBER OF EXTRACTED SAMPLES FROM EACH SUBJECT

| Subject ID | File Name | Sample Number | |
|---|---|---|---|
| | | *Alert* | *Drowsy* |
| 1 | s01_061102n.set | 94 | 94 |
| 2 | s05_061101n.set | 66 | 66 |
| 3 | s22_090825n.set | 75 | 75 |
| 4 | s31_061103n.set | 74 | 74 |
| 5 | s35_070322n.set | 112 | 112 |
| 6 | s41_080520m.set | 83 | 83 |
| 7 | s42_070105n.set | 51 | 51 |
| 8 | s43_070205n.set | 132 | 132 |
| 9 | s44_070325n.set | 157 | 157 |
| 10 | s45_070307n.set | 54 | 54 |
| 11 | s53_090918n.set | 113 | 113 |
| | Total | 1011 | 1011 |

From (1), we can observe that the LSTM function takes an element $x_t$ at the time step $t \in 1 ... T$ and the previous hidden state $h_{t-1}$ as input and returns a new hidden state $h_t$. In this way, each $h_t$ can be viewed as an encryption of essential information from $(x_1, x_2, ..., x_t)$. The last hidden vector $h_T$ is expected to contain all the essential information from the whole input signal that is useful for classification. $h_T$ is usually linked to a dense layer in the network structure, e.g., [19], for summarizing the features.

Although it is widely acknowledged that the hidden-state sequence $(h_1, h_2, ..., h_T)$ contains important information for predicting the label of the input sample, it is still difficult to decrypt the information from the vectors [20]. In order to make the hidden states interpretable, we have come up with a novel way of reducing the hidden dimension $D$ of LSTM as the number of classes (which is 2 in our case) and directly linking $h_T$ to the final classification of the network by a Softmax activation layer. The proposed method not only largely reduces the number of parameters and encourages convergence of the network, but also makes $(h_1, h_2, ..., h_T)$ to show how the prediction by the network changes in the process of adding new information incrementally to the model. The overall model structure can be seen in Fig. 1.

#### 2) Model structure

In the model structure, CNN is used to extract a feature set $X$ from the input signal, which is followed by an LSTM layer to link the extracted features to the final classification. The CNN structure we use is similar to the one proposed in our previous work [12]. The whole model consists of six layers, which are a convolutional layer, a batch normalization layer, an ELU activation layer, an average pooling layer, an LSTM layer and a Softmax Activation layer.

- The convolutional layer contains 32 one-dimensional convolutional kernels that shift on the input signal with the stride of 1. Each kernel has the length of 64, which is half the signal sampling rate (128Hz), so that it can capture essential EEG frequencies above 2Hz. The input samples are padded along the temporal dimension in order to keep the output from this layer the same length as the input signal.
- The Batch Normalization (BN) operation is used in the second layer, in order to remove internal covariate shift.
- The Exponential Linear Unit (ELU) function is used in the activation layer to add non-linearity transformation to the data.
- Average pooling operation with a window size of (1, 8) is used to summarize the extracted features. The output of the layer is a feature set of $(x_1, x_2, \ldots, x_T)$, $x_i \in \mathbb{R}^K$, where $T(=48)$ is length of the temporal dimensional and $K(=32)$ is the size of each feature.
- The LSTM layer maps the feature sequence to hidden representations $(h_1, h_2, \ldots, h_T)$, $h_i \in \mathbb{R}^D$, where $D(=2)$ is the dimension of each representation.
- The final activation layer performs Softmax function on $h_T$ to output the classification results.

*3) Visualization technique*

Visualization on what features from EEG signals have been learned by deep learning models is an important topic, since it can reveal whether biologically explainable patterns rather than noise and artifacts have been used by the model for classification. In our work, we want to design a visualization technique that can identify the local regions of an input signal that contribute most to the classification. This will allow us to understand what micro-structures nested in EEG signals have been learned by the network with relation to different mental states and to compare the findings with existing known neurophysiological patterns of EEG signals.

Specifically, suppose an EEG sample $S = (s_1, s_2 \ldots, s_n)$ is classified with label $c$ by the network. In our case, $c=0$ or 1, representing the alert label or the drowsy label. We want to generate a heatmap $M_c = (m_1, m_2 \ldots, m_n)$, where each $m_i$ is a value indicating the importance of the sampling point $s_i$ to the final classified label $c$. In the area of image processing, special techniques have been proposed to extract $M_c$ from deep CNN models, e.g., the Class Activation Map (CAM) [21]. For our model that is specially designed for ease of interpretation, $M_c$ can be directly generated from the hidden-state sequence $(h_1, h_2, \ldots, h_T)$.

According to the discussion in Section III.B.2), the final activation can be expressed as

$$h_T^* = Softmax(h_T) \quad (2)$$

$h_T^*$ is a 2-dimensional vector that contains the likelihood of both classes output by the network. We apply the Softmax function to each hidden state $h_t$

$$h_t^* = Softmax(h_t) \quad (3)$$

In this way, we can get a new set of vectors $(h_1^*, h_2^*, \ldots, h_T^*)$ that reflect how the likelihood changes when the LSTM model is incrementally updated with $x_t$. Suppose the likelihood of class $c$ at the time $t$ is $h_{t,c}^*$, the sequence $H_c^* = (h_{1,c}^*, h_{2,c}^*, \ldots, h_{T,c}^*)$ reflects the evolution process of the how the network reaches the final likelihood $h_{T,c}^*$, or in another perspective, how each $x_t$ subtly modifies the likelihood of class $c$. The contribution of $x_t$ to the final classification of the network can be therefore evaluated by

$$\Delta h_{t,c}^* = h_{t,c}^* - h_{t-1,c}^*, \quad (4)$$

where $t \in 1 \ldots T$ and $h_{0,c}^* = 0$. Finally, we calculate $\widetilde{\Delta H_c^*} = (\widetilde{\Delta h_{1,c}^*}, \widetilde{\Delta h_{2,c}^*}, \ldots, \widetilde{\Delta h_{T,c}^*})$ by normalizing $\Delta H_c^* = (\Delta h_{1,c}^*, \Delta h_{2,c}^*, \ldots, \Delta h_{T,c}^*)$ (mean removal of $\Delta H_c^*$ followed by dividing each element with the standard deviation). In fact, $\widetilde{\Delta H_c^*}$ is similar to the class activation map in the method [21].

Finally, we consider how to get the final heatmap $M_c$ from $\widetilde{\Delta H_c^*}$. By observing the network structure, we can find that the first three layers do not change the length (temporal dimension) of the input data. Only the fourth average pooling layer reduce the temporal dimension of the data to 1/8 of the original size. Therefore, a direct way to get $M_c$ from $\widetilde{\Delta H_c^*}$ is padding the elements eight times to reverse the average pooling operation so that it can reach the desired length. Therefore, $M_c$ can be obtained from

$$M_c = (\underbrace{\widetilde{\Delta h_{1,c}^*}, \ldots, \widetilde{\Delta h_{1,c}^*}}_{8}, \ldots, \underbrace{\widetilde{\Delta h_{T,c}^*}, \ldots, \widetilde{\Delta h_{T,c}^*}}_{8}) \quad (5)$$

Similarly, we can get an additional map $M_c^*$ from the original likelihood evolution sequence $H_c^*$ by

$$M_c^* = (\underbrace{h_{1,c}^*, \ldots, h_{1,c}^*}_{8}, \ldots, \underbrace{h_{T,c}^*, \ldots, h_{T,c}^*}_{8}) \quad (6)$$

In this way, we can have obtained two maps $M_c$ and $M_c^*$ from the hidden states. In order to distinguish them, we name $M_c$ as the "relative heatmap", since it reflects the relative importance of local regions of the input signal to the final classification. We name $M_c^*$ as the "accumulated heatmap", since it reflects the accumulated contribution of local areas to the likelihood of class $c$.

*C. Methods for comparison*

*1) Deep learning methods*

We implement three state-of-art deep learning models for comparison. The first model for comparison is the Deep CNN model proposed by Fahimi et al. [9], which was used for distinguishing between the attention and non-attention states from single-channel EEG signals. Since the original network is designed for dealing with EEG signals of 85.33Hz, we modify the kernel size of the first convolutional layer from (1, 4) to (1, 6) and change the stride from (1, 6) to (1, 3), in order to adapt to our data. All the other parameters are set to be the same as that described in the original paper.

The second deep learning model we implement for comparison is the EEGNet model proposed by Lawhern et al. [22] for classifying multi-channel EEG signals. In our implementation, we remove the spatial convolutional layer and the following batch normalization layer in the original model, since they were designed for multi-channel EEG signals. We set the number of filters in the first convolutional layer as $F_1 = 8$ in our study.

The third deep learning model we use for comparison is the Compact CNN model [12], which was proposed by Cui et al. previously for the same purpose of interpretable cross-subject driver drowsiness recognition. The model uses a GAP layer in the structure, so that it can localize regions of the input signal that contribute most for classification with the CAM method. The source code of the model is accessible from [23].

*2) Conventional baseline methods*

We implement three conventional methods for feature extraction and test them with different classifiers for comparison.

**RelativePower**: relative powers of frequency bands Delta (1–4 Hz), Theta (4–8 Hz), Alpha (8–12 Hz) and Beta (12–30Hz) are calculated from the Oz channel with the Welch's method to form the feature set.

**PowerRatio**: we implement the method proposed by Jap et al. [24], where four band power ratios (i) $(θ + α)/β$, (ii) $α/β$, (iii) $(θ + α)/(α + β)$ and (iv) $θ/β$ are used for driver drowsiness recognition.

**FourEntropies**: In the work by Hu et al. [25], four entropy features, which are sample entropy, fuzzy entropy, approximate entropy and spectral entropy, were used for driver fatigue recognition. We implement the method according to the description in the paper. Specifically, the approximate entropy and spectral entropy features are extracted with the methods proposed by Song et al. [26], while extraction of the fuzzy entropy was based on the method proposed by Xiang et al. [27]. We set the entropy parameters m = 2 and r = 0.2 * SD, and the width of the exponential function n involved in fuzzy entropy as n=2.

**Classifiers**: The methods above are tested with several commonly used classifiers, including Decision Tree (DT), Random Forest (RF), k-nearest neighbors (KNeighbors), Gaussian Naive Bayes (GNB), Logistic Regression (LR), Linear Discriminant Analysis (LDA), Quadratic Discriminant Analysis (QDA), and Support Vector Machine (SVM).

*D. Implementation details*

We implement and run the codes in the environment of Python 3.6.6 on a Windows 10 platform powered by Intel(R) Core(TM) i7-6700 CPU and an NVIDIA GeForce GTX 1080 graphics card. Leave-one-subject-out cross validation is performed on the dataset. Specifically, each time the data from one subject is used for testing, while the data from all the other subjects are used for training. In this way, we can get a classification accuracy for each subject. For the neural network models, we use Adam method [28] with parameters of $η = 0.01$, $β_1 = 0.9$ and $β_2 = 0.999$ for training. The classification accuracies are recorded from 1 to 50 epochs. We repeat the process by 10 times and the mean accuracy for each epoch is reported. The source code of the model and visualization technique is accessible from [29].

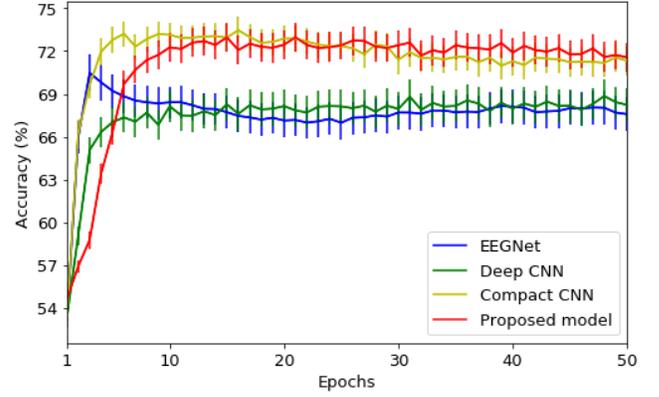

Figure 2. Comparison on the mean classification accuracies (%) of Deep CNN, EEGNet and the proposed model against training epochs from 1 to 50.

## IV. EVALUATION ON THE PROPOSED METHOD

*A. Model comparison results*

From Fig. 2, we can see that the proposed model and Compact CNN have higher accuracies than the other deep learning baseline models. The best accuracy 72.97% of the proposed model is achieved after 15 training epochs. The Compact CNN model converges faster and reached a similar accuracy around 73%. In comparison, the best mean accuracy of EEGNet is 70.48%, which is achieved after 3 training epochs, while the best accuracy of Deep CNN in the first 50 training epochs is 68.85%, which is achieved after 48 training epochs. It can be inferred from the obtained results that the proposed model and Compact CNN have a better capability in identifying shared features of EEG signals across different subjects for drowsiness recognition than the other two models.

As it can be seen in TABLE II, the mean accuracies obtained with conventional baseline methods of RelativePower, PowerRatio and FourEntropies range from 55.42%-69.27%, which are lower than the best mean accuracies achieved by the proposed model (72.97%). The best accuracies for all the three conventional baseline methods of RelativePower, PowerRatio and FourEntropies, are achieved with the SVM classifier, which are 69.27%, 66.64% and 63.36%, respectively. The method of RelativePower has the highest mean accuracy of 69.27% with different classifiers, which is followed by the methods of PowerRaio (66.64%) and FourEntropies (63.36%).

Finally, we compare the proposed model with the three baseline methods on the accuracies for each subject. For the three conventional baseline methods, we obtain the results with the SVM classifier, since it achieves the highest mean accuracies with all the three methods. For the proposed model, we obtain the mean accuracy for each subject by averaging the results of 10 repetitions after 6 training epochs. The results are shown in TABLE III. As it can be seen in the table, around half (5 out of 11) of the highest individual accuracies are achieved by the proposed model. Paired t-tests show that the proposed model has the mean accuracy significantly higher than RelativePower (p=0.014), PowerRatio (p=0.008) and FourEntropies (p=0.012).

TABLE II. MEAN CLASSIFICATION ACCURACIES (%) AND THE STANDARD DEVIATION (SD) OF THE THREE CONVENTIONAL BASELINE MEHTODS COMBINED WITH DIFFERENT CLASSIFIERS

| Classifiers | RelativePower (SD) | PowerRatio (SD) | FourEntropies (SD) |
|---|---|---|---|
| DT | 60.70 (8.64) | 56.49 (6.16) | 55.99 (5.63) |
| RF | 62.06 (10.97) | 59.27 (8.80) | 58.67 (6.60) |
| KNeighbors | 63.42 (10.67) | 62.23 (8.45) | 55.59 (8.73) |
| GaussianNB | 67.44 (13.44) | 59.47 (8.67) | 59.40 (10.64) |
| LR | 67.77 (13.54) | 65.42 (13.32) | 59.80 (9.71) |
| LDA | 67.60 (15.03) | 65.68 (13.1) | 59.62 (9.57) |
| QDA | 67.08 (14.52) | 56.97 (9.41) | 60.80 (10.35) |
| SVM | **69.27 (13.38)** | **66.64 (13.06)** | **63.36 (10.48)** |
| Mean | 65.67 | 61.71 | 59.05 |

TABLE III. COMPARISON ON THE MEAN ACCURACIES (%) OF THE PROPOSED MODEL AND THE THREE CONVENTIONAL METHODS WITH THE SVM CLASSIFIER FOR EACH SUBJECT. THE OBTAINED MEAN ACCURACIES OF THE PROPOSED MODEL ARE OBTAINED AFTER 15 TRAINING EPOCHS.

| Subject ID | Relative Power | Power Ratio | Four Entropies | Proposed Model |
|---|---|---|---|---|
| 1 | 77.13 | 76.60 | **79.79** | 78.35 |
| 2 | 46.21 | 40.91 | **55.30** | 52.05 |
| 3 | 49.33 | 54.00 | **63.33** | 59.07 |
| 4 | 61.49 | 57.43 | 45.95 | **70.27** |
| 5 | 68.75 | 59.38 | 49.11 | **74.11** |
| 6 | **85.54** | 81.93 | 69.88 | 83.43 |
| 7 | 63.73 | 61.76 | 52.94 | **67.55** |
| 8 | 73.48 | **78.79** | 66.67 | 71.89 |
| 9 | 81.21 | 78.34 | 71.34 | **88.31** |
| 10 | **86.11** | 82.41 | 74.07 | 85.19 |
| 11 | 69.03 | 61.50 | 68.58 | **72.48** |
| Mean (SD) | 69.27 (13.38) | 66.64 (13.70) | 63.36 (10.99) | **72.97** (10.92) |

## B. Visualization on the learned characteristics from EEG signals

Having obtained the results, we consider visualizing what features from EEG signals have been learned by the model as indicators of alertness and drowsiness across different subjects. Specifically, we get the accumulated heatmap ($M_c^*$ in Section III.B.3) and the relative heatmap ($M_c$ in Section III.B.3) with the proposed technique. The accumulated heatmap displays the evolution of the likelihood when the input signal is incrementally visible to the LSTM model. From the accumulated heatmap we are able to know to which extent each local region of the signal contributes to the likelihood of the output class $c$. The relative heatmap is obtained by applying normalization on the difference of the accumulated heatmap so that it reflects which parts of the input signal contribute most to the classification.

We display two samples classified as alertness correctly by the model with a high likelihood in Fig. 3(a) and (b), and another two samples classified correctly as drowsiness by the model with a high likelihood in Fig. 3(c) and (d). The accumulated heatmap and the relative heatmap are indicated by the color maps of the input sample, and they are shown in the left column of each sub-figure. In addition, we also display the ratio of wave components in Delta, Theta, Alpha and Beta frequency bands of the signal, as well as the EEG signals from all the other channels in the sampling period as reference.

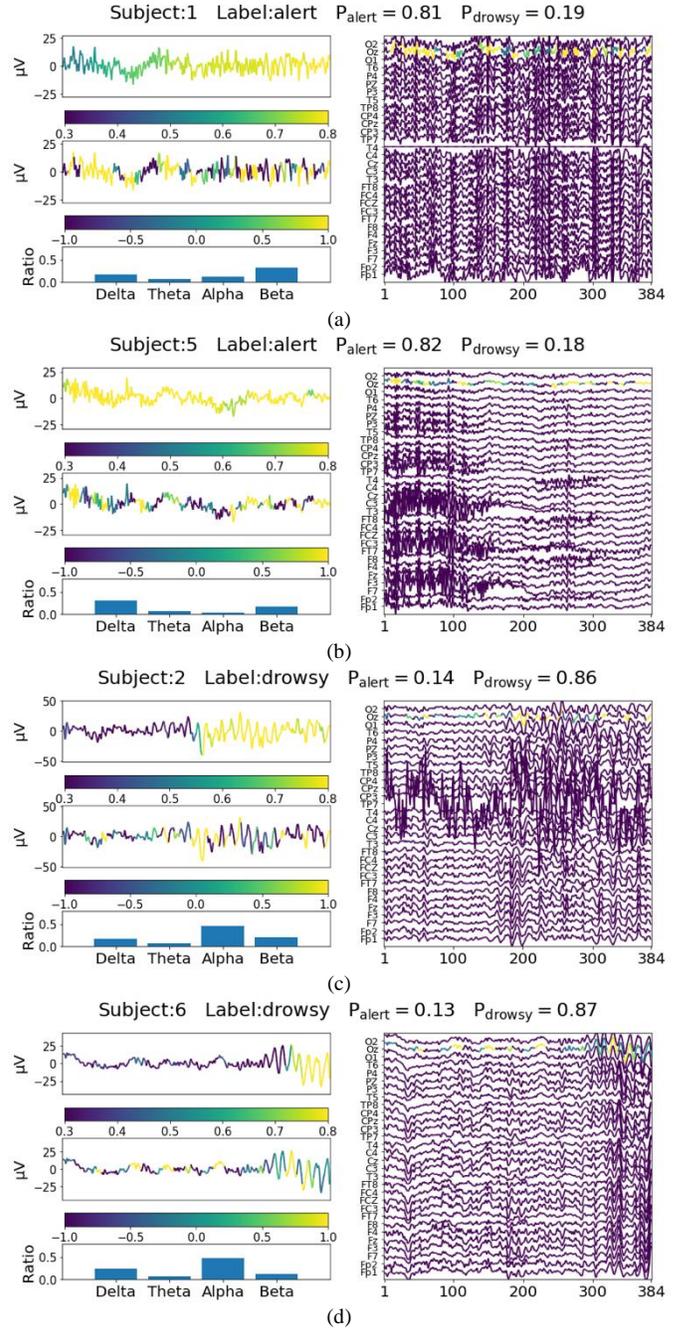

Figure 3. Visualization results for selected samples classified correctly with high likelihood of the alert label (a), (b) and fatigue label (c), (d). The subject ID, label, likelihood for each class are displayed for each sample. The accumulated heatmap $M_c^*$ and the relative heatmap $M_c$ are shown on the top and middle of the left column, respectively, in each sub-figure. The relative powers on Delta, Theta, Alpha and Beta frequency bands for each sample signal from Oz channel are shown on the bottom of each left column, and the "whole picture" of each sample with EEG signals from all the 30 channels (including Oz) in the sampling period is shown on the right part of each sub-figure.

For the samples classified with a high likelihood of alertness, we have found that they commonly contain a high portion of

Beta waves, e.g., Fig. 3(a) and 3(b). Some also contain a high portion of Delta waves, e.g., Fig. 3(b). For the first sample shown in Fig. 3(a), we find from the relative heatmap that the model has localized several parts in the signal that contain a high portion of fast waves as evidence of alertness. The accumulated heatmap shows that the likelihood of the alert class is stably increasing when the input signal is gradually sent to the LSTM model, which indicates that the highlighted local part in the relative heatmap contributes incrementally to the likelihood of the alert class. The pattern of relative heatmap for the second sample shown in Fig. 3(b) is similar to that of the first one, which indicates that similar types of features have been recognized by the model as evidence of the alert state. However, the accumulated heatmap of the second sample is different from the first one in the aspect it has high likelihood (above 0.8) of the alert class throughout the sample capturing period. This indicates that the model has identified strong alertness-related features from the beginning parts of the sample.

By observing the "whole picture" of the first sample with signals from all the channels, we can find that the EEG data contains heavy high-frequency noise in almost all of the channels. Actually, the noise is caused by electromyography (EMG) activities that usually dominate the wakeful EEG signals [30] and often have peaks in the Beta frequency range that resemble EEG Beta peaks [31]. For the second sample shown in Fig. 3(b), it can be observed that the similar type of EMG noise contaminates the signals but mostly of the beginning part, which lasts for around 1 second (corresponding to around 150 sampling points). The EMG noise is centered at the frontal-central area of the scalp with the strongest amplitude and it propagates to all the other channels. In addition, it is also inferred from the multi-channel signals that the large portion of Delta waves contained in the second sample is generated by sensor drifts or displacements associated with subject movements, which causes fluctuations of signals simultaneously in all the channels and they are clearly visible from the channels of F7, FT7 and T3. Therefore, it can be concluded that high likelihood of the alert state for the samples are majorly caused by EMG activities and sensor displacements caused by subject movements, which usually dominates the wakeful EEG signals.

We have found that a large number of samples classified with high likelihood of the drowsy state contains spindle-like structures that cause a high power in the Alpha frequency band, e.g., samples shown in Fig. 3(c) and 3(d). For the sample shown in Fig. 3(c), it can be observed from the accumulated heatmap that the classification is majorly contributed by features detected from the second half of the signal (the sensor noise observable from TP7 does not affect the classification), while for the last sample shown in Fig. 3(d) the classification is majorly contributed by the end region of the sample. The corresponding multi-channel EEG signals show that the spindle-like structures also appear in the EEG signals from the other channels. Actually, the captured micro-structures in EEG signals are named "Alpha spindles", which can be characterized by a narrow frequency peak within the alpha band [32]. Existing works have found that these structures are strong indicators of early drowsiness and they have been used to identify drowsiness in various driving simulator studies [33].

## V. DISCUSSION AND FUTURE WORK

In this paper, we consider the task of subject-independent driver drowsiness recognition from EEG signals. In order to discover common features of EEG signals shared by different subjects, we designed a LSTM-CNN model that can explain the features and reveal which part of the data contribute most to the classification. The model uses CNN to extract features and LSTM to learn the temporal dependency of the feature sequence. The learned features from the input samples can be visualized from the hidden states output by the LSTM layer.

With the proposed model, we have discovered interesting features from the data that are associated with the alert and the drowsy states from the diverse EEG signals of different subjects. Specifically, the model has discovered neurologically interpretable "Alpha spindles" features as indicators of the drowsy state, while it identifies EMG activities and artefacts associated with subject movements as indicators of the alert state. The obtained initial results are encouraging – it shows that there indeed exists such common features among different subjects that can be used for subject-independent drowsiness recognition. It could be a promising direction to use interpretable deep learning techniques for subject-independent mental state recognition and the classification accuracy can be improved in the following aspects in our future works.

**Data cleaning**. In fact, it is out of our expectation that the EMG activities, which are conventionally regarded as artefacts and usually suppressed in the pre-processing stage, could be beneficial for the classification task. It implicates that a cleaning pipeline should be specially designed for a specific classification task rather than using the same cleaning routine for every task. The interpretable deep learning can be incorporated in the process to guide the design of the cleaning pipeline, so that features contained in the signals that are beneficial for classification can be maximally retained, while those that are harmful for classification, e.g., sensor noise, should be removed in the cleaning stage.

**Model comparison**. In order to design a model with the best performance, sample-wise comparison can be conducted on different candidate models to find out reasons that cause the difference of performance. In this way, we will be able to finally get the model with the optimal structure and parameters that can best capture shared features from the data.

## VI. CONCLUSION

In this paper, we proposed a novel CNN-LSTM model for subject-independent drowsiness recognition from single-channel EEG signals. The network uses CNN to extract features from the data and use LSTM to link the extracted features to the final output of the network. In order to reveal what features have been learned by the model from the input samples, we designed a novel way that makes hidden states output by the LSTM model directly show how local regions of the signal contributes to the final classification. Results show that the model achieves an average leave-one-subject-out accuracy of 72.97% on 11 subjects, which is higher than both the conventional baseline methods and state-of-the-art deep learning methods. Visualization results show that the model has identified meaning EEG patterns, e.g., Alpha spindles, to distinguish between alert and drowsy mental states. Our work shows a promising direction of EEG-based subject-independent mental state recognition with interpretable deep learning models.

ACKNOWLEDGEMENT

This research is supported by the National Research Foundation, Singapore under its International Research Centres in Singapore Funding Initiative. Any opinions, findings and conclusions or recommendations expressed in this material are those of the author(s) and do not reflect the views of National Research Foundation, Singapore.